\documentclass[11pt]{article}
\pdfoutput=1
\usepackage[utf8]{inputenc}
\usepackage{acl2016}
\usepackage{times}
\usepackage{latexsym}
\usepackage{mathrsfs} 
\usepackage{amsmath} 
\usepackage{amssymb} 
\usepackage{mathtools} 
\usepackage{xparse}
\usepackage{enumitem} 
\usepackage[normalem]{ulem} 
\usepackage[scaled=.86]{helvet} 

\newcommand{\cev}[1]{\reflectbox{\ensuremath{\vec{\reflectbox{\ensuremath{#1}}}}}}

\aclfinalcopy 

\title{Robust Named Entity Recognition in Idiosyncratic Domains}

\author{Sebastian Arnold\qquad Felix A. Gers\qquad Torsten Kilias\qquad Alexander Löser\\
  Beuth University of Applied Sciences \\
  Luxemburger Straße 10 \\
  13353 Berlin, Germany \\
  {\tt \{sarnold,gers,tkilias,aloeser\}@beuth-hochschule.de}
\\}

\date{}

\begin{document}

\maketitle

\begin{abstract}
Named entity recognition often fails in idiosyncratic domains. That causes a problem for depending tasks, such as entity linking and relation extraction. We propose a generic and robust approach for high-recall named entity recognition. Our approach is easy to train and offers strong generalization over diverse domain-specific language, such as news documents (e.g. Reuters) or biomedical text (e.g. Medline). Our approach is based on deep contextual sequence learning and utilizes stacked bidirectional LSTM networks. Our model is trained with only few hundred labeled sentences and does not rely on further external knowledge. We report from our results F1 scores in the range of 84–94\% on standard datasets.
\end{abstract}

\section{Introduction}

Information extraction tasks have become very important not only in the Web, but also for in-house enterprise settings. One of the crucial steps towards understanding natural language is named entity recognition (NER), which aims to extract mentions of entity names in text. NER is necessary for many higher-level tasks such as entity linking, relation extraction, building knowledge graphs, question answering and intent based search. In these scenarios, NER recall is critical, as candidates that are never generated can not be recovered later \cite{hachey2013evaluating}.

\paragraph{Challenges.} \newcite{pink2014analysing} show that NER components can reduce the search space for slot filling tasks by 99.8\% with a recall loss of 15\%. However, large effort is required to adapt most annotators to specialized domains, such as biomedical documents. When focusing on recall for these domains, we face three major problems. First, the language used in the documents is often idiosyncratic and cannot be effectively identified by standard natural language processing (NLP) tools \cite{prokofyev2014effective}. 
Second, training these domains is difficult: data is sparse, data may contain a large number of non-linkable entity mentions (NILs) and large labeled gold standards are hardly available. 
Third, applications vary greatly and we cannot standardize annotation guidelines to meet all of their requirements \cite{ling2015design}. For example, NER on news texts might focus on proper named entity annotation (e.g. people, companies and locations), whereas phrase recognition on medical text might include the annotation of common concepts (e.g. medical terms and treatments).
We therefore focus on a generalized NER component with high recall, which can be trained ad-hoc with only few labeled examples.

\paragraph{Common error analysis.}
\newcite{ling2015design} point out common errors of NER systems, which yield non-recognized mentions (false negatives), invalid detections (false positives), wrong boundaries (e.g. multi-word mentions, missing determiners) and annotation errors from human labelers (e.g., correct answers are not marked as correct, unclear annotation guidelines).
Consider the following example taken from the biomedical GENIA corpus \cite{ohta2002genia}, with underlined named entity mentions:
\begin{quote}
\textsf{\textbf{Example:}
Engagement of the \uline{Lewis X antigen (CD15)} results in \uline{monocyte activation}.
Nuclear extracts of anti-\uline{CD15 cross-linked cells} demonstrated enhanced levels of the \uline{transcriptional factor activator protein-1}, minimally changed \uline{nuclear factor-kappa B}, and did not affect \uline{SV40 promoter specific protein-1}.
}
\end{quote}
We observe that common errors originate from a manifold number of sources, which are frequently:
\begin{itemize}[noitemsep]
\item non-verbatim mentions (e.g. misspellings, alternate writings: \textsf{\uline{monoyctes}}, \textsf{\uline{Lewis-X}})
\item part-of-speech (POS) tagging errors (e.g. unidentified NP tags: \textsf{\uline{monoycte}/JJ})
\item wrong capitalization (e.g. uppercase headlines, lowercase proper names)
\item unseen or novel words (e.g. idiosyncratic language: \textsf{anti-CD15})
\item irregular word context (e.g. collapsed lists, semi-structured data, invalid segmentation)
\end{itemize}

\paragraph{Our contribution.} We contribute DATEXIS-NER, a generic annotator for robust named entity recognition that can be trained for various domains with low human labeling effort. DATEXIS-NER does not depend on domain-specific rules, dictionaries, fine-tuning, syntactic annotation or external knowledge bases. Instead, our approach is built from scratch and is based on core character features of text. We train our model for news and biomedical domains with raw text data and few hundred labels. From our results, we report equal performance compared to state-of-the-art NER annotators with high 90\% F1 scores for common NER corpora, such as CoNLL2003, KORE50, ACE2004 and MSNBC. We show on the highly domain-specific biomedical GENIA corpus that our approach adapts to various idiosyncratic domains. In particular, we observe that bidirectional long short-term memory (LSTM) networks capture useful distributional context for NER applications and generic letter-trigram word encoding with surface forms compensates typing and capitalization errors. With a combination of these techniques, we achieve better context representation than word2vec models trained with significantly larger corpora. 

The rest of this paper is structured as follows. In Section \ref{sec:relatedwork}, we discuss related work. We introduce our approach of robust named entity recognition in Section \ref{sec:methodology}. In Section \ref{sec:evaluation}, we evaluate our approach compared to state-of-the-art annotators and discuss the most common errors in the components of our system. We conclude in Section \ref{sec:summary} and propose future work on our approach.

\section{Related Work}
\label{sec:relatedwork}

\paragraph{Named entity recognition.} The task of NER has been extensively studied with various evaluation in the last decades: MUC-6, MUC-7, CoNLL2002, CoNLL2003 and ACE. The standard approach to NER is the application of discriminative tagging \cite{collins2002discriminative} to the task of NER \cite{mccallum2003early}, often with linear chain Conditional Random Field (CRF), Hidden Markov (HMM) or Maximum Entropy Hidden Markov Models (MEMM). Later, \newcite{bengio2003neural} used continuous-space language models, where type-to-vector word mappings can be learned using backpropagation. \newcite{mikolov2013efficient} achieved a more effective vector representation using the skip-gram model. The model optimizes the likelihood of tokens over a window surrounding a given token. This training process produces a linear classifier that predicts words conditioned on the central token's vector representation. 

\paragraph{Recall bounds for idiosyncratic entity linking.} Named entity linking is the task to match textual mentions of named entities to a knowledge base \cite{shen2015entity}. This task requires a set of candidate mentions from sentences. As a result, the recall from the underlying NER system constitutes an upper bound for entity linking accuracy \cite{hachey2013evaluating}. Moreover, \newcite{pink2014analysing} show that ``state-of-the-art systems are substantially limited by low recall'' and don't perform well especially on idiosyncratic data while \newcite{prokofyev2014effective} highlight that terms with high novelty or high specificity cannot efficiently be linked by current systems.

\paragraph{State-of-the-art NER implementations.} We distinguish between three broad categories for generating candidate entities: Babelfy \cite{moro2014entity}, Entityclassifier.eu \cite{dojchinovski2013entityclassifier}, DBpedia Spotlight \cite{mendes2011dbpedia} or TagMe2 \cite{ferragina2010tagme} spot noun chunks and filter them with dictionaries, often derived from Wikipedia. Stanford NER \cite{manning2014stanford} or LingPipe\footnote{http://alias-i.com/lingpipe/} utilize discriminative tagging approaches. FOX \cite{speck2014ensemble} or NERD-ML \cite{vanerp2013learning} combine several approaches in an ensemble learner for enhancing precision. The GENIA tagger\footnote{http://www.nactem.ac.uk/tsujii/GENIA/tagger/} is a tagger specifically tuned for biomedical text.
It is trained on the GENIA-based BioNLP/NLPBA 2004 data set \cite{kim2004introduction} that includes named entity recognition for biomedical text.
The biomedical NER system of \newcite{zhou2004exploring} is built using HMM and an additional SVM with sigmoid. It uses lexical-level features, e.g. word formation and morphological patterns, and utilizes dictionaries. 
The system of \newcite{finkel2004exploiting} uses a MEMM.
\newcite{settles2004biomedical} use CRF classifiers with syntactical features and synset dictionaries.
Basically, all these systems benefit from our work.

\section{Robust Contextual Word Labeling}
\label{sec:methodology}

\begin{figure}[t!]
\centering{\includegraphics[clip=false,width=1.0\columnwidth]{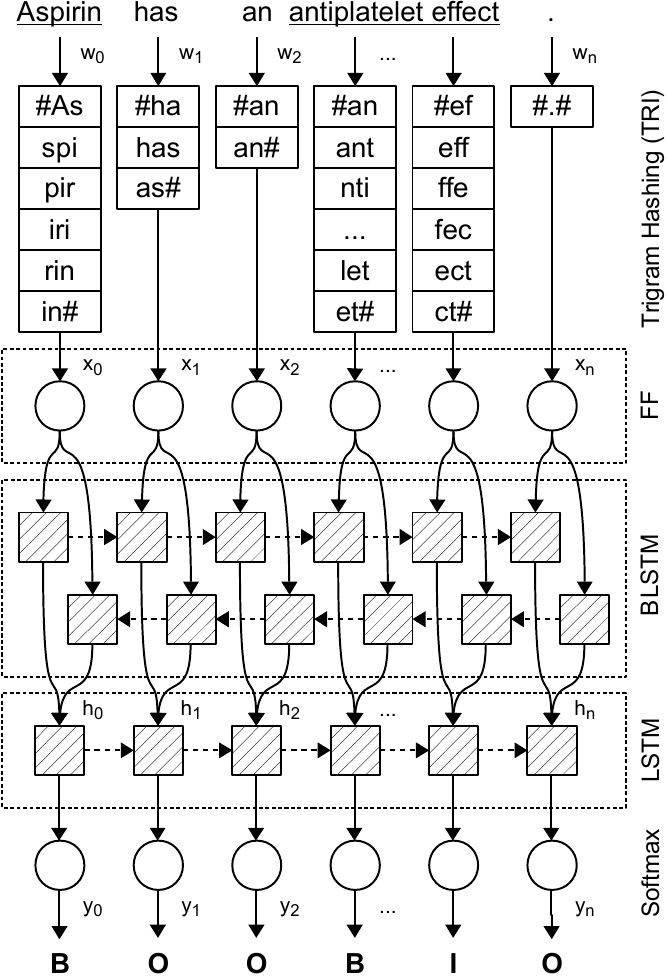}}
\caption{\label{fig:lstm} Architecture of the LSTM network used for named entity recognition. The character stream ``\textsf{\uline{Aspirin} has an \uline{antiplatelet effect}.}'' is tokenized into words $w_t$ and converted into word vectors $x_t$ using letter-trigram hashing. These vectors are propagated through a recurrent neural network with four layers: (1) feed-forward encoding of word vectors, (2+3) bidirectional LSTM layers for context representation using forward and backward passes, (4) LSTM decoder layer for context-sensitive label prediction. The output labels $y_t$ follow the BIO2 standard and represent mention begin (B), inside (I) and outside (O) per token.}
\end{figure}

We abstract the task of NER as sequential word labeling problem. Figure \ref{fig:lstm} illustrates an example for sequential transformation of a sentence into word labels. We express each sentence in a document as a sequence of words: $w=(w_0,w_1,\dots,w_n)$, e.g. $w_0 =$ \textsf{Aspirin}. We define a mention as the longest possible span of adjacent tokens that refer to a an entity or relevant concept of a real-world object, such as \textsf{\uline{Aspirin (ASA)}}. We further assume that mentions are non-recursive and non-overlapping. To encode boundaries of the mention span, we adapt the idea of \newcite{ramshaw1995text}, which has been adapted as BIO2 standard in the CoNLL2003 shared task \cite{tjongkimsang2003introduction}. We assign labels $\{B,I,O\}$ to each token to mark begin, inside and outside of a mention from left to right. We use the input sequence $w$ together with a target sequence $y$ of the same length that contains a BIO2 label for each word: $y=(y_0,y_1,\dots,y_n)$, e.g. $y_0 =$ B. To predict the most likely label $\hat{y}_t$ of a token regarding its context, we utilize recurrent neural networks.

\subsection{Robust Word Encoding Methods}
\label{sec:encoder}

We have shown that most common errors for recall loss are misspellings, POS errors, capitalization, unseen words and irregular context. Therefore we generalize our model throughout three layers: robust word encoding, in-sentence word context and contextual sequence labeling.

\paragraph{Letter-trigram word hashing to overcome spelling errors.} Dictionary-based word vectorization methods suffer from sparse training sets, especially in the case of non-verbatim mentions, rare words, typing and capitalization errors. For example, the word2vec model of \newcite{mikolov2013efficient} generalizes insufficiently for rare words in idiosyncratic domains or for misspelled words, since for these words no vector representation is learned at training time. In the GENIA data set, we notice 27\% unseen words (dictionary misses) in the pretrained word2vec model\footnote{GoogleNews-vectors-negative300 embeddings (3.6 GB)}. As training data generation is expensive, we investigate a generic approach for the generation of word vectors. We use letter-trigram word hashing as introduced by \newcite{huang2013learning}. This technique goes beyond words and generates word vectors as a composite of discriminative three-letter ``syllables'', that might also include misspellings. Therefore, it is robust against dictionary misses and has the advantage (despite its name) to group syntactically similar words in similar vector spaces. We compare this approach to word embedding models such as word2vec. 

\paragraph{Surface form features for word vector robustness.} The most important features for NER are word shape properties, such as length, initial capitalization, all-word uppercase, in-word capitalization and use of numbers or punctuation \cite{ling2012fine}. Mixed-case word encodings implicitly include capitalization features. However, this approach impedes generalization, as words appear in various surface forms, e.g. capitalized at the beginning of sentences, uppercase in headlines, lowercase in social media text. The strong coherence between uppercase and lowercase characters – they might have identical semantics – is not encoded in the embedding. Therefore, we encode the words using lowercase letter-trigrams. To keep the surface information, we add flag bits to the vector that indicate initial capitalization, uppercase, lower case or mixed case.

\subsection{Deep Contextual Sequence Learning}
\label{sec:decoder}

With sparse training data in the idiosyncratic domain, we expect input data with high variance. Therefore, we require a strong generalization for the syntactic and semantic representation of language. To reach into the high 80–90\% NER F1 performance, long-range context-sensitive information is indispensable. We apply the computational model of recurrent neural networks, in particular long short-term memory networks (LSTMs) \cite{hochreiter1997long,gers2002learning} to the problem of sequence labeling. Like neural feed-forward networks, LSTMs are able to learn complex parameters using gradient descent, but include additional recurrent connections between cells to influence weight updates over adjacent time steps. With their ability to memorize and forget over time, LSTMs have proven to generalize context-sensitive sequential data well \cite{graves2012supervised,lipton2015critical}.

Figure \ref{fig:lstm} shows an unfolded representation of the steps through a sentence. We feed the LSTM with letter-trigram vectors $x_t$ as input data, one word at a time. The hidden layer of the LSTM represents context from long range dependencies over the entire sentence from left to right. However, to achieve deeper contextual understanding over the boundaries of multi-word annotations and at the beginning of sentences, we require a backwards pass through the sentence. We therefore implement a bidirectional LSTM and feed the output of both directions into a second LSTM layer for combined label prediction.

\paragraph{Bidirectional sequence learning.} For the use in the neural network, word encodings $x_t$ and labels $y_t$ are real-valued vectors. To predict the most likely label $\hat{y}_t$ of a token, we utilize a LSTM with input nodes $g_t$, input gates $i_t$, forget gate $f_t$, output gate $o_t$ and internal state $s_t$. For the bidirectional case, all gates are duplicated and combined into forward state $h_t$ and backward state $z_t$. The network is trained using backpropagation through time (BPTT) by adapting weights $W$ and bias parameters $b$ to fit the training examples.
\begin{align}
\begin{split}
g_t &= \phi(W_{gx} x_t + W_{gh} h_{t-1} + b_g)\\
i_t &= \sigma(W_{ix} x_t + W_{ih} h_{t-1} + b_i)\\
f_t &= \sigma(W_{fx} x_t + W_{fh} h_{t-1} + b_f)\\
o_t &= \sigma(W_{ox} x_t + W_{oh} h_{t-1} + b_o)\\
s_t &= \phi(g_t \odot i_t + s_{t-1} \odot f_t)\\
h_t &= \vec{s_t} \odot \vec{o_t} \quad/\quad z_t = \cev{s_t} \odot \cev{o_t}\\
y_t &= \mathit{softmax}(W_{yh} h_t + W_{yz} z_t + b_y)
\end{split}
\end{align} 
We iterate over labeled sentences in mini-batches and update the weights accordingly. The network is then used to predict label probabilities $y_t$ for unseen word sequences $w_t$.

\subsection{Implementation of NER Components \label{sec:configurations}}

To show the impact of our bidirectional LSTM model, we measure annotation performance on three different neural network configurations. We implement all components using the Deeplearning4j framework\footnote{http://deeplearning4j.org, version 0.4-rc3.9-SNAPSHOT}. For preprocessing (sentence and word tokenization), we use Stanford CoreNLP\footnote{version 3.6.0} \cite{manning2014stanford}. We test the sequence labeler using three input encodings:
\begin{itemize}[noitemsep]
\item DICT: We build a dictionary over all words in the corpus and generate the input vector using 1-hot encoding for each word
\item EMB: We use the GoogleNews word2vec embeddings, which encodes each word as vector of size 300
\item TRI: we implement letter-trigram word hashing as described in Section \ref{sec:encoder}.
\end{itemize}
During training and test, we group all tokens of a sentence as mini-batch. We evaluate three different neural network types to show the impact of the bidirectional sequence learner.
\begin{itemize}[noitemsep]
\item FF: As baseline, we train a non-sequential feed-forward model based on a fully connected multilayer perceptron network with 3 hidden layers of size 150 with relu activation, feeding into a 3-class softmax classifier. We train the model using backpropagation with stochastic gradient descent and a learning rate of 0.005.
\item LSTM: We use a configuration of a single feed-forward layer of size 150 with two additional layers of single-direction LSTM with 20 cells and a 3-class softmax classifier. We train the model using backpropagation-through-time (BPTT) with stochastic gradient descent and a learning rate of 0.005.
\item BLSTM: Our final configuration consists of a single feed-forward layer of size 150 with one bidirectional LSTM layer with 20 cells and an additional single-direction LSTM with 20 cells into a 3-class softmax classifier. The BLSTM model is trained the same way as the single-direction LSTM.
\end{itemize}

\section{Evaluation}
\label{sec:evaluation}

We evaluate nine configurations of our model on five gold standard evaluation data sets. We show that the combination of letter-trigram word hashing with bidirectional LSTM yields the best results and outperforms sequence learners based on dictionaries or word2vec. To highlight the generalization of our model to idiosyncratic domains, we run tests on common-typed data sets as well as on specialized medical documents. We compare our system on these data sets with specialized state-of-the-art systems.

\subsection{Evaluation Set Up}

We train two models with identical parameterization, each with 2000 randomly chosen labeled sentences from a standard data set. To show the effectiveness of the components, we evaluate different configurations of this setting with 2000 random sentences from the remaining set. The model was trained using Deeplearning4j with nd4j-x86 backend. Training the TRI+BLSTM configuration on a commodity Intel i7 notebook with 4 cores at 2.8GHz takes approximately 50 minutes.

\paragraph{Evaluation data sets.} Table \ref{tab:datasets} gives an overview of the standard data sets we use for training. The GENIA Corpus \cite{ohta2002genia} contains biomedical abstracts from the PubMed database. We use GENIA technical term annotations 3.02, which cover linguistic expressions to entities of interest in molecular biology, e.g. proteins, genes and cells. CoNLL2003 \cite{kim2004introduction} is a standard NER dataset based on the Reuters RCV-1 news corpus. It covers named entities of type person, location, organization and misc.

For testing the overall annotation performance, we utilize CoNLL2003-testA and a 50 document split from GENIA. Additionally, we test on the complete KORE50 \cite{hoffart2012kore}, ACE2004 \cite{mitchell2005ace} and MSNBC data sets using the GERBIL evaluation framework \cite{usbeck2015gerbil}.

\begin{table}[t]
\centering{\includegraphics[clip=false,width=1.0\columnwidth]{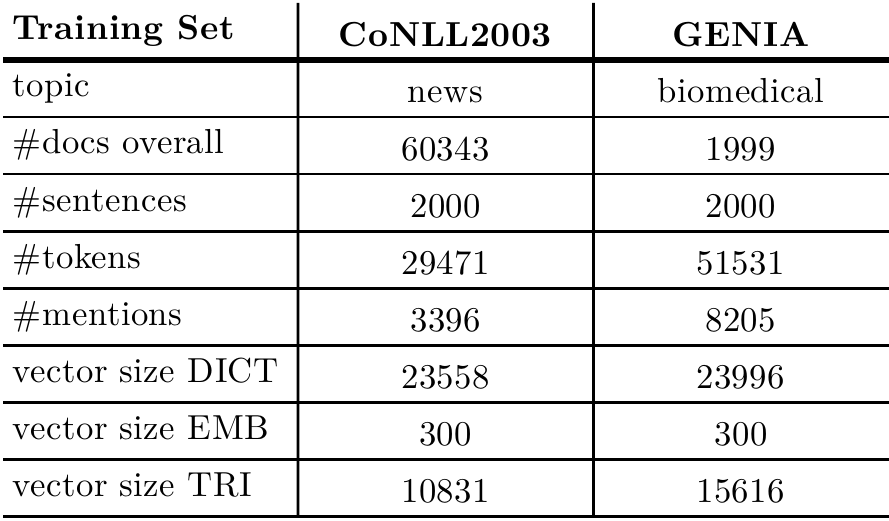}}
\caption{\label{tab:datasets} Overview of CoNLL2003 and GENIA training datasets and sizes of word encodings. We use 2000 sentences of each set for training.}
\end{table}

\subsection{Measurements}

We measure precision, recall and F1 score of our DATEXIS-NER system and state-of-the-art annotators introduced in Section \ref{sec:relatedwork}. For the comparison with black box systems, we evaluate annotation results using weak annotation match. For a more detailed in-system error analysis, we measure BIO2 labeling performance based on each token.

\begin{table*}[th!]
\centering{\includegraphics[clip=false,width=1.0\textwidth]{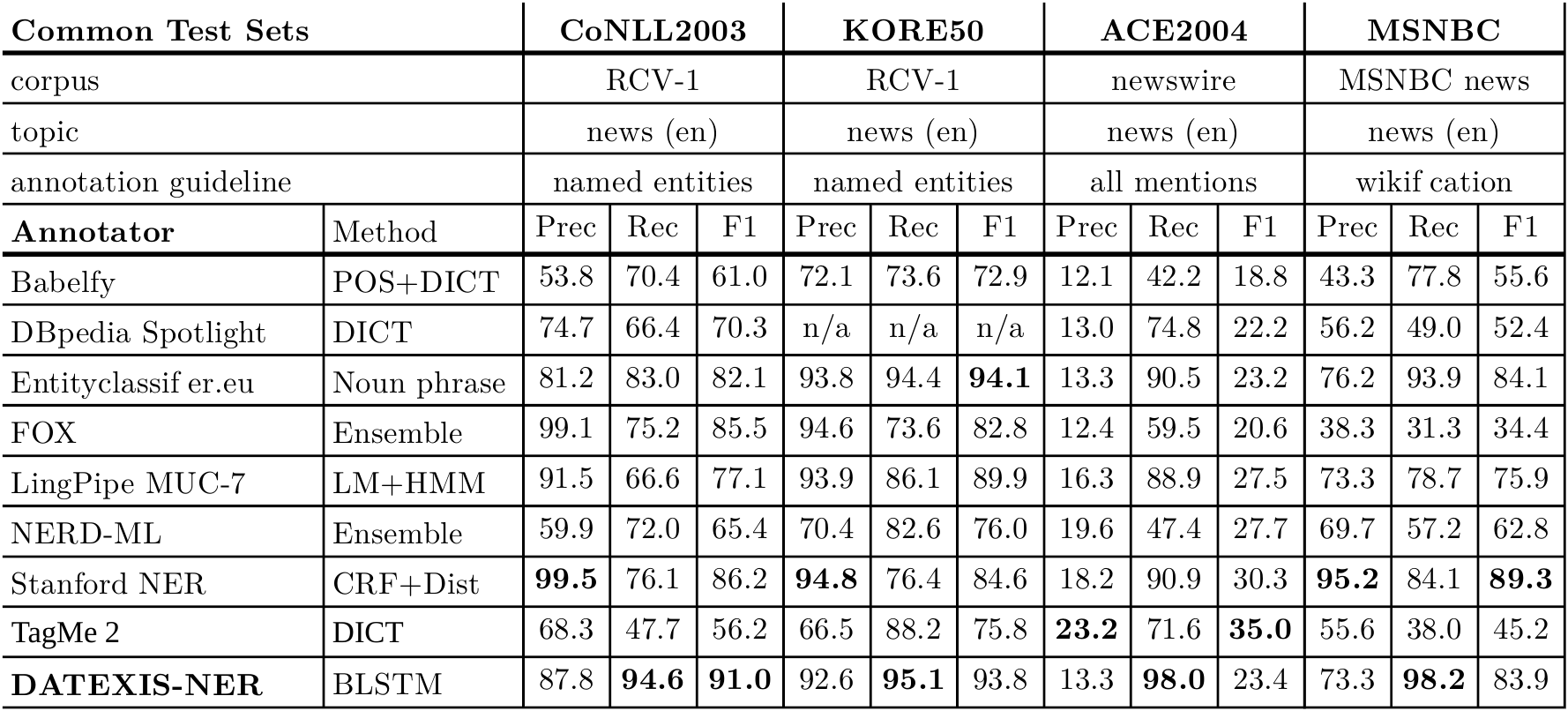}}
\caption{\label{tab:comparison} Comparison of annotators trained for common English news texts (micro-averaged scores on match per annotation span). The table shows micro-precision, recall and NER-style F1 for CoNLL2003, KORE50, ACE2004 and MSNBC datasets.
}
\end{table*}

\paragraph{Measuring annotation performance using NER-style F1.} We measure the overall performance of mention annotation using the evaluation measures defined by \newcite{cornolti2013framework}, which are also used by \newcite{ling2015design}. Let $\mathcal{D}$ be a set of documents with gold standard mention annotations $\mathcal{G} = \{\mathcal{G}_d \mid d \in \mathcal{D}\}$ with a total of $N = |\mathcal{G}|$ examples. Each mention $g_i \in \mathcal{G}$ is defined by start position $b$ and end position $e$ in the source document $d$. To quantify the performance of the system, we compare $\mathcal{G}$ to the set of predicted annotations $\mathcal{P} = \{\mathcal{P}_d \mid d \in \mathcal{D}\}$ with mentions $p_i \in \mathcal{P}$:
\begin{align}
\begin{split}
\mathit{tp_d} &= \lvert\{p \in \mathcal{P}_d \mid \exists g \in \mathcal{G}_d : \mathit{m}(p,g)\}\rvert\\
\mathit{fp_d} &= \lvert\{p \in \mathcal{P}_d \mid \nexists g \in \mathcal{G}_d : \mathit{m}(p,g)\}\rvert\\
\mathit{tn_d} &= \lvert\{p \notin \mathcal{P}_d \mid \nexists g \in \mathcal{G}_d : \mathit{m}(p,g)\}\rvert\\
\mathit{fn_d} &= \lvert\{g \in \mathcal{G}_d \mid \nexists p \in \mathcal{P}_d : \mathit{m}(g,p)\}\rvert
\end{split}
\end{align}
We compare using a weak annotation match:
\begin{align}
\begin{split}
\mathit{m}: (p,g) \mapsto & (b_p \leq b_g \leq e_p) \vee  \\
& (b_p \leq e_g \leq e_p) \vee  \\
& (b_g \leq b_p \leq e_g) \vee  \\
& (b_q \leq e_p \leq e_g)
\end{split}
\end{align}
We measure micro-averaged 
precision ($Prec$), recall ($Rec$) and NER-style ($\mathit{F1}$) score:
\begin{align}
\begin{split}
\mathit{Prec} &= \frac{\sum_{d \in \mathcal{D}}\mathit{tp_d}}{\sum_{d \in \mathcal{D}}\left(\mathit{tp_d} + \mathit{fp_d}\right)}\\
\mathit{Rec} &= \frac{\sum_{d \in \mathcal{D}}\mathit{tp_d}}{\sum_{d \in \mathcal{D}}\left(\mathit{tp_d} + \mathit{fn_d}\right)}\\
\mathit{F1} &= \frac{2\cdot\mathit{Prec}\cdot\mathit{Rec}}{\mathit{Prec}+\mathit{Rec}}
\end{split}
\end{align}

\paragraph{Measuring BIO2 labeling performance.} Tuning the model configuration with annotation match measurement is not always feasible. We therefore measure $tp_c$, $fp_c$, $tn_c$, $fn_c$ separately for each label class $c \in \{B,I,O\}$ in our classification model and calculate binary classification precision $Prec_c$, recall $Rec_c$ and  $\mathit{F1}_c$ scores. To avoid skewed results from the expectedly large $O$ class, we use macro-averaging over the three classes:
\begin{align}
\begin{split}
\mathit{Prec}_{BIO} &= \frac{1}{3}\sum_{c \in \{B,I,O\}} \frac{tp_c}{tp_c+fp_c}\\
\mathit{Rec}_{BIO} &= \frac{1}{3}\sum_{c \in \{B,I,O\}} \frac{tp_c}{tp_c+fn_c}
\end{split}
\end{align}

\subsection{Evaluation Results}

We now discuss the evaluation of our DATEXIS-NER system on common and idiosyncratic data.

\paragraph{Overall model performance on common types.} Table \ref{tab:comparison} shows the comparison of DATEXIS-NER with eight state-of-the-art annotators on four common news data sets. Both common and medical models are configured identically and trained on only 2000 labeled sentences, without any external prior knowledge. We observe that DATEXIS-NER achieves the highest recall scores of all tested annotators, with 95\%–98\% on all measured data sets.  Moreover, DATEXIS-NER precision scores are equal or better than median. Overall, we achieve high micro-F1 scores of 84\%–94\% on news entity recognition, which is slightly better than the ontology-based Entityclassifier.eu NER and reveals a better generalization than the 3-type Stanford NER with distributional semantics. We notice that systems specialized on word-sense disambiguation (Babelfy, DBpedia Spotlight) don't perform well on ``raw'' untyped entity recognition tasks. The highest precision scores are reached by Stanford NER. We also notice a low precision of all annotators on the ACE2004 dataset and high variance in MSNBC performance, which are probably caused by differing annotation standards.

\paragraph{Biomedical recognition performance.} Table \ref{tab:medical} shows the results of biomedical entity recognition compared to the participants of the JNLPBA 2004 bio-entity recognition task \cite{kim2004introduction}. We notice that for these well-written Medline abstracts, there is not such a strong skew between precision and recall. Our DATEXIS-NER system outperforms the HMM, MEMM, CRF and CDN based models with a micro-F1 score of 84\%. However, the highly specialized GENIA chunker for LingPipe achieves higher scores. This chunker is a very simple generative model predictor that is based on a sliding window of two tokens, word shape and dictionaries. We interpret this score as strong overfitting using a dictionary of the well-defined GENIA terms. Therefore, this model will generalize hardly considering the simple model. We can confirm this presumption in the common data sets, where the MUC-6 trained HMM LingPipe chunker performs on average on unseen data.

\begin{table}[t]
\centering{\includegraphics[clip=false,width=1.0\columnwidth]{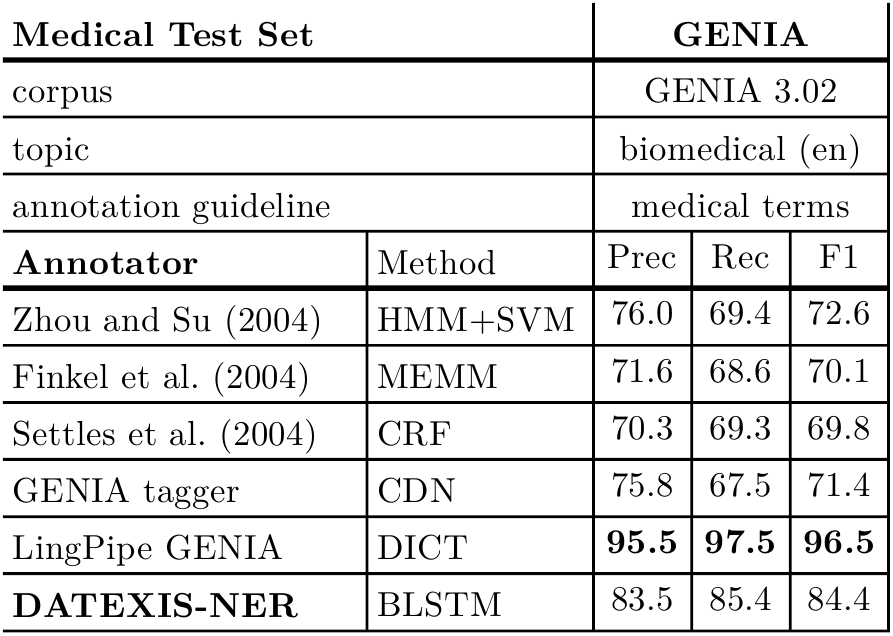}}
\caption{\label{tab:medical} Comparison of annotators trained for biomedical text. The table shows NER annotation results for 50 documents from the GENIA dataset.}
\end{table}

\paragraph{Evaluation of system components.} We evaluate different configurations of the components that we describe in Section \ref{sec:configurations}. Table \ref{tab:configurations} shows the results of experiments on both CoNLL2003 and GENIA data sets. We report the highest macro-F1 scores for BIO2 labeling for the configuration of letter-trigram word vectors and bidirectional LSTM. We notice that dictionary-based word encodings (DICT) work well for idiosyncratic medical domains, whereas they suffer from high word ambiguity in the news texts. Pretrained word2vec embeddings (EMB) perform well on news data, but cannot adapt to the medical domain  without retraining, because of a large number of unseen words. Therefore, word2vec generally achieves a high precision on news texts, but low recall on medical text. The letter-trigram approach (TRI) combines both word vector generalization and robustness towards idiosyncratic language. 

We observe that the contextual LSTM model achieves scores throughout in the 85\%–94\% range and significantly outperforms the feed-forward (FF) baseline that shows a maximum of 75\%. Bidirectional LSTMs can further improve label classification in both precision and recall.

\begin{table}[bt]
\centering{\includegraphics[clip=false,width=1.0\columnwidth]{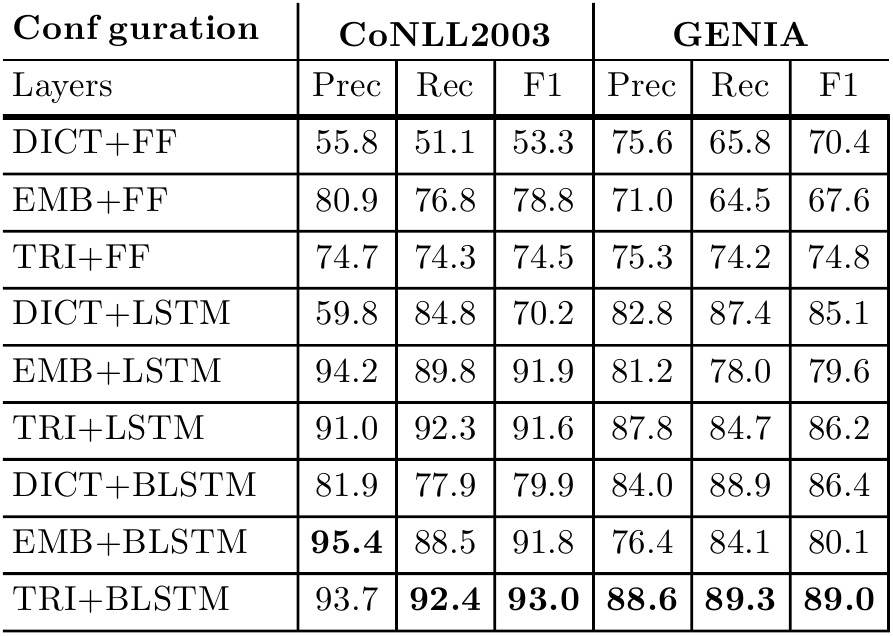}}
\caption{\label{tab:configurations} Comparison of nine configurations from our implementation (macro-averaged scores on BIO2 classification per token).}
\end{table}

\subsection{Discussion and Error Analysis}

We investigate different aspects of the DATEXIS-NER components by manual inspection of classification errors in the context of the document. For the error classes described in the introduction (false negative detections, false positives and invalid boundaries), we observe following causes:

\paragraph{Unseen words and misspellings.}  In dictionary based configurations (e.g. 1-hot word vector encoding DICT), we observe false negative predictions caused by dictionary misses for words that do not exist in the training data. The cause can be rare unseen or novel words (e.g. \textsf{\uline{T-prolymphocytic cells}}) or misspellings (e.g. \textsf{strengthnend}). These words yield a null vector result from the encoder and can therefore not be distinguished by the LSTM. The error increases when using word2vec, because these models are trained with stop words filtered out. This implicates that e.g. mentions surrounded by or containing a determiner (e.g. \textsf{\uline{The Sunday Telegraph} quoted \uline{Majorie Orr}}) are highly error prone towards the detection of their boundaries. We resolve this error by the letter-trigram approach. Unseen trigrams (e.g. \textsf{thh}) may still be missing in the word vector, but only affect single dimensions as opposed to the vector as a whole.

\paragraph{Misleading surface form features.} Surface forms encode important features for NER (e.g. capitalization of ``new'' in \textsf{\uline{Alan Shearer} was named as the new \uline{England} captain} / \textsf{as \uline{New York} beat the \uline{Angels}}). However, case-sensitive word vectorization methods yield a large amount of false positive predictions caused by incorrect capitalization in the input data. An uppercase headline (e.g. \textsf{TENNIS - \uline{U.S.} TEAM ON THE ROAD FOR \uline{1997 FED CUP}}) is encoded completely different than a lowercase one (e.g. \textsf{\uline{U.S.} team on the road for \uline{Fed Cup}}). Because of that, we achieve best results with lowercase word vectors and additional surface form feature flags, as described in Section \ref{sec:encoder}.



\paragraph{Syntagmatic word relations.} We observe mentions that are composed of co-occurring words with high ambiguity (e.g. \textsf{degradation of \uline{IkB alpha} in \uline{T cell lines}}). These groups encode strong syntagmatic word relations \cite{sahlgren2008distributional} that can be leveraged to resolve word sense and homonyms from sentence context. Therefore, correct boundaries in these groups can effectively be identified only with contextual models such as LSTMs.

\paragraph{Paradigmatic word relations.} Orthogonal to the previous problem, different words in a paradigmatic relation \cite{sahlgren2008distributional} can occur in the same context (e.g. \textsf{\uline{cyclosporin A-}treated cells} / \textsf{\uline{HU} treated cells}). These groups are efficiently represented in word2vec. However, letter-trigram vectors cannot encode paradigmatic groups and therefore require a larger training sample to capture these relations.

\paragraph{Context boundaries.} Often, synonyms can only be resolved regarding a larger document context than the local sentence context known by the LSTM. In these cases, word sense is redefined by a topic model local to the paragraph (e.g. sports: \textsf{\uline{Tiger} was lost in the woods after divorce.}). This problem does not heavily affect NER recall, but is crucial for named entity disambiguation and coreference resolution. 

\paragraph{Limitations.} The proposed DATEXIS-NER model is restricted to recognize boundaries of generic mentions in text. We evaluate the model on annotations of isolated types (e.g. persons, organizations, locations) for comparison purposes only, but we do not approach NER-style typing. Contrary, we approach to detect mentions without type information. The detection of specific types can be realized by training multiple independent models on a selection of labels per type and nesting the resulting annotations using a longest-span semantic type heuristic \cite{kholghi2015external}.

\section{Summary}
\label{sec:summary}

\newcite{ling2015design} show that the task of NER is not clearly defined and rather depends on a specific problem context. Contrary, most NER approaches are specifically trained on fixed datasets in a batch mode. Worse, they often suffer from poor recall \cite{pink2014analysing}. Ideally, one could personalize the task of recognizing named entities, concepts or phrases according to the specific problem. ``Personalizing'' and adapting such annotators should happen with very limited human labeling effort, in particular for idiosyncratic domains with sparse training data.

Our work follows this line. From our results we report F1 scores between 84–94\% when using bidirectional multi-layered LSTMs, letter-trigram word hashing and surface form features on only few hundred training examples.

This work is only a preliminary step towards the vision of personalizing annotation guidelines for NER \cite{ling2015design}. In our future work, we will focus on additional important idiosyncratic domains, such as health, life science, fashion, engineering or automotive. For these domains, we will consider the process of detecting mentions and linking them to an ontology as a joint task and we will investigate simple and interactive workflows for creating robust personalized named entity linking systems.

\paragraph{Acknowledgements}
Our work is funded by the Federal Ministry of Economic Affairs and Energy (BMWi) under grant agreement 01MD15010B (Project: Smart Data Web).

\bibliography{sarnold2016idiosyncratic.bib}
\bibliographystyle{acl2016}
\end{document}